%% file: main.tex
\documentclass[10pt,twocolumn,letterpaper]{article}

\usepackage{cvpr_include/cvpr,times,graphicx,tabulary,multirow,xspace,xcolor}
\usepackage{amsmath, amssymb}
\usepackage{amsthm,fixmath,mathtools,nicefrac}
\usepackage{subfig}
\usepackage{epsfig}
\usepackage{booktabs}
\usepackage[
    pagebackref=false,breaklinks=true,letterpaper=true,
    colorlinks,citecolor=citecolor,
    bookmarks=false
]{hyperref}
\definecolor{citecolor}{RGB}{0,113,188}

\cvprfinalcopy 
\def\cvprPaperID{} 

\newif\ifshowcomment

\newcolumntype{x}[1]{
    >{\centering\arraybackslash}p{#1pt}
}
\newcolumntype{y}[1]{
    >{\raggedright\arraybackslash}p{#1pt}
}
\newcolumntype{z}[1]{
    >{\raggedleft\arraybackslash}p{#1pt}
}
\newlength\savewidth\newcommand\shline{
    \noalign{\global\savewidth\arrayrulewidth\global\arrayrulewidth 1pt}
    \hline\noalign{\global\arrayrulewidth\savewidth}
}
\newcommand{\tablestyle}[2]{
    \setlength{\tabcolsep}{#1}
    \renewcommand{\arraystretch}{#2}
    \centering
    \footnotesize
}

\newcommand{\vf}[0]{\emph{Voting Filter}\xspace}
\newcommand{\qf}[0]{\emph{Query Filter}\xspace}
\newcommand{\pr}[0]{\emph{peak region}\xspace}
\def\argmax{\mathop{\rm argmax}}

\begin{document}
\title{Pixel Consensus Voting for Panoptic Segmentation}

\author{
Haochen Wang \thanks{Work done when Haochen was at UChicago and working at TTI-C}\\
Carnegie Mellon University \\
{\tt\small whcw@cmu.edu}
\and
Ruotian Luo\\
TTI-Chicago\\
{\tt\small rluo@ttic.edu}
\and
Michael Maire\\
University of Chicago\\
{\tt\small mmaire@uchicago.edu}
\and
Greg Shakhnarovich\\
TTI-Chicago\\
{\tt\small greg@ttic.edu}
}

\maketitle

\begin{abstract}
 \input{abs}
\end{abstract}

\input{intro}

\input{related}

\input{pcv-desc}

\input{exps}

\input{conclusion}

\section{Acknowledgements}
We would like to thank Deva Ramanan for discussions and feedback.
The work was in part supported by the DARPA L2M award
FA8750-18-2-0126, the DARPA
GARD award HR00112020003,  and AFOSR award FF9950-18-1-0166 (MADlab).

{\small \bibliographystyle{cvpr_include/ieee_fullname} \bibliography{references}}

\end{document}

%% file: abs.tex
The core of our approach, Pixel Consensus Voting, is a framework for instance segmentation
based on the Generalized Hough transform. Pixels cast discretized, probabilistic votes
for the likely regions that contain instance centroids.
At the detected peaks that emerge in the voting heatmap,
backprojection is applied to collect pixels and produce instance masks.
Unlike a sliding window detector that densely enumerates object proposals,
our method detects instances as a result of the consensus among pixel-wise votes.
We implement vote aggregation and backprojection using native operators of a convolutional neural network.
The discretization of centroid voting reduces the training of instance segmentation to pixel labeling,
analogous and complementary to FCN-style semantic segmentation,
leading to an efficient and unified architecture that jointly models things and stuff.
We demonstrate the effectiveness of our pipeline on COCO and Cityscapes Panoptic Segmentation
and obtain competitive results. Code will be open-sourced.

%% file: intro.tex

\vspace{-1em}
\section{Introduction}\label{sec:intro}

\begin{figure*}[!t]
\centering
\includegraphics[width=1\linewidth]{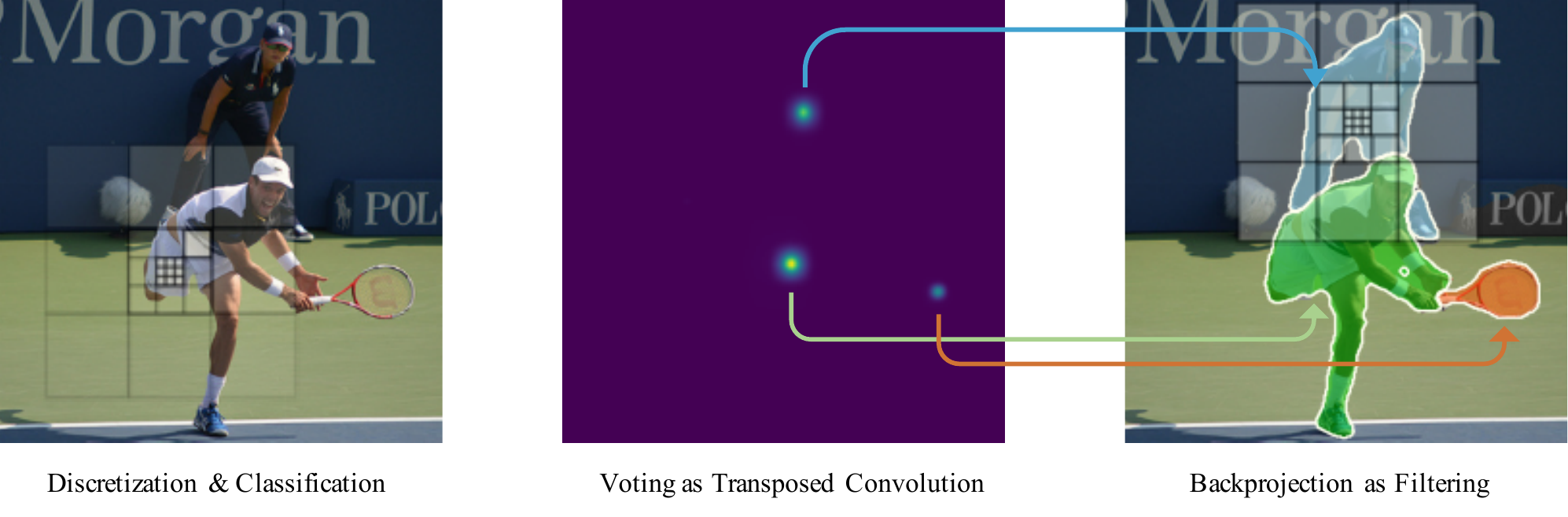}
\vspace{-1em}
\caption{\small
   An overview of PCV.
   \textbf{Left}: A large region around each pixel is discretized into spatial cells according
   to the \vf. The size of the cells expands the farther they are from
   the pixel. A convnet votes for the location of
   the instance to which the pixel belongs, in the form of a probability distribution 
   over the cells that might contain the instance centroid; 
   a pixel can also vote for ``abstaining'' (not belonging to any object).
   \textbf{Middle}: The votes from each pixel are aggregated
   into a voting heatmap, using dilated deconvolution
   (transposed conv). \emph{Peak Regions} of the heatmap are
   treated as initial instance detection hypotheses.
   \textbf{Right}: \qf, the spatial inversion of the \vf, is convolved with each \pr to 
   backproject an instance mask for that peak.
   Not shown: the semantic segmentation branch assigns categories to
   things and stuff. See also Fig.~\ref{fig:insfig}.}
\label{fig:teaser}\vspace{-3mm}
\end{figure*}

The development of visual recognition algorithms has followed the evolution of
recognition benchmarks. PASCAL VOC~\cite{pascal} standardizes the task of
bounding box object detection and the associated IoU/Average Precision metrics.
At the time, the approaches defining the state-of-the-art, DPM~\cite{DPM} and
later the R-CNN family~\cite{rcnn,frcnn}, address object detection by
reasoning about densely enumerated box proposals, following the sliding window
classification approach of earlier detectors~\cite{viola_jones, earlydet1}.
SDS~\cite{hariharan2014simultaneous} expands the scope of object detection to
include instance mask segmentation, and introduces early versions of
mAP$^{\text{bbox}} $ and mAP$^{\text{mask}} $,
subsequently popularized by the COCO dataset~\cite{mscoco}.
Bounding boxes, however, remain the primary vehicle for object
reasoning.

The more recently introduced task of panoptic segmentation~\cite{PStask} removes the
notion of boxes altogether. It treats both ``things'' and ``stuff'' in
a unified format, in which ground truth and predictions are expressed as
labelled segment masks: instance and category for objects (``things''), and
category only for ``stuff''~\cite{coco-stuff}.

Our work focuses on this particular framework of image understanding. We propose
an approach, Pixel Consensus Voting (PCV), that in line with this updated task
definition, elevates \emph{pixels} to first class citizen status.
Every pixel contributes evidence for the presence, identity, and location of
an object to which it may belong. PCV aggregates and backprojects this evidence,
in a Hough transform-like framework, so that detections emerge from the
\emph{consensus} among pixels that vote for consistent object hypotheses.
Figure~\ref{fig:teaser} summarizes our approach.

Notably, and in contrast to the currently dominant line of detection work
derived from the R-CNN family~\cite{rcnn, frcnn, mrcnn,tmask}, our approach
does not involve reasoning about bounding boxes.
It can be seen as a descendant of early Hough-based methods such as the
Implicit Shape Model (ISM)~\cite{ISM, leibe2008robust}, traditionally
referred to as ``bottom-up''.
We extend these earlier works and leverage the
power of convolutional networks for feature extraction. Since these
features have large receptive fields and presumably capture high-level
semantic concepts, it is unclear whether the bottom-up designation remains
appropriate. Another distinction from prior attempts to use voting is our
vote representation. Traditional methods treat voting as offset regression, and
suffers from the problem of ``regressing to the mean", where prediction is conflated with
uncertainty. In PCV, we treat voting for object location as classification over discretized
spatial cells. This allows for the representation of uncertainty and for
``abstention'' votes by non-object pixels.
It also lends itself to efficient vote aggregation and backprojection using
(dilated \cite{dilated_conv}) convolutional mechanisms.

Despite its simplicity, PCV achieves competitive results on COCO
and Cityscapes panoptic segmentation benchmarks.
On COCO, PCV outperforms all existing
proposal-free or single-stage detection methods. Our work revisits a classic
idea from a modern perspective, and we expect that future research extending
our approach will yield major performance gains and novel insights.

%% file: related.tex
\section{Related work}\label{sec:related}

\paragraph{Sliding window detection}
Over the last two decades, most approaches to object detection and instance
segmentation have followed the general pipeline by ranking sliding window
proposals. In a typical setup, a large number of candidate regions are
sampled from an input image and a predictor (usually a convolutional network)
scores each region's likelihood to intersect with objects. For highly ranked
proposals, the network also predicts their categories, bounding box
coordinates, and optionally generates instance masks. Two stage methods, such
as Faster/Mask R-CNN~\cite{frcnn,mrcnn} use regional feature pooling as an
attention mechanism to enhance prediction accuracy, while single
stage methods, including YOLO~\cite{yolo}, RetinaNet~\cite{retinanet} and
SSD~\cite{SSD}, combine all network decisions in a single feedforward pass.
Another line of work eschews object proposals and predicts keypoint
heatmap as a proxy for object localization
~\cite{law2018cornernet, zhou2019objects, tian2019fcos, zhou2020tracking}. 
\vspace{-1em}

\paragraph{Instances from pixels}
Many attempts have been made to establish a direct connection between image
pixels and instance segmentations. A natural idea to group pixels into
instances is to obtain some measure of pixel affinity for clustering.
In~\cite{disembedding, newell-asso}, a network is trained to produce pixel
embeddings that are similar within and different across instances, and an
off-the-shelf clustering algorithm is used for grouping.
RPE~\cite{kong2018recurrent} integrates the clustering step into the learning
process by formulating mean-shift clustering as a recurrent neural network.
AdaptIS~\cite{adapis} further improves the training of discriminative
embeddings by a novel scheme that provides end-to-end supervision.
In addition, exploiting the sparsity in pairwise pixel affinity makes it possible
to instantiate the clustering step as a \mbox{graph-cut}.
Learning the sparse pixel affinity can be formulated as boundary
detection~\cite{kirillov2017instancecut}, or a richer multi-hop dilated connectivity prediction~\cite{ssap,liu2018affinity}.
Alternatively, instances may be generated sequentially using a recurrent neural network
~\cite{ren2017end, torr-recurrent}, or treated as watershed basins through a learned boundary
distance transform ~\cite{watershed_bai}.

\paragraph{Detections via Generalized Hough transform}
The Hough transform~\cite{hough} frames the task of detecting analytical shapes
as identifying peaks in a dual parametric space; this idea can be
generalized~\cite{ghough} to arbitrary objects. The gist of the Generalized Hough
transform is to collect local evidence as \emph{votes} for the likely
location, scale, and pose of potential instances. 
Works in this vein such as Implicit Shape Models~\cite{ISM} rely on memorized
mapping from image patches to offsets, and is later improved by the use of
more advanced learning techniques~\cite{majihough,houghforests}. It has been
applied to a variety of problems including
pose estimation~\cite{houghpose, houghposelets},
tracking~\cite{houghforests} and 3D object detection~\cite{3dhough}.
Our work can be seen as a descendant of these earlier efforts.

Some recent works follow broadly similar philosophy, but differ from ours in
many ways. Most~\cite{neven_clustering, personlab, deeperlab} treat learning to
vote for object centroids as a regression task. Our work avoids the potential limitations
of offset regression and uses discretized region classification to capture
pixel-level uncertainty. We design convolutional mechanisms for efficient vote
aggregation and backprojection under this classification setting.

To our knowledge, the only prior work using transposed convolution \ie deconv  for
classification-based pixel voting is~\cite{pose_voting}, applied to single
person pose estimation. We take inspiration from their work, but differ in
motivation and implementation.

\paragraph{Panoptic Segmentation}
Most existing work address panoptic segmentation by merging
the outputs from specialized components designed for instance~\cite{mrcnn}
and semantic segmentation~\cite{pspnet, deeplab} with greedy
heuristics~\cite{PStask}. PFPN \cite{PFPN} establishes a strong single network
baseline by sharing the FPN~\cite{FPN} feature for Mask R-CNN~\cite{mrcnn} and
FCN~\cite{FCN} sub-branches.
\cite{upsnet, tascnet, li2019attention, liu2019end} improve the segment
overlap resolution with learned modules. \cite{SSPS, fastps} trade off the
performance for speed by using single-stage object detectors. Proposal-free
methods~\cite{adapis, ssap, deeperlab, pdeeplab} provide novel perspectives
that directly model instances from pixels, but in general lag behind the
performance of those leveraging mature engineering solutions, with an
especially large gap on the challenging COCO~\cite{mscoco} benchmark.

%% file: pcv-desc.tex
\section{Pixel Consensus Voting}\label{sec:pcv-desc}

\begin{figure}[ht]
    \centering
    \includegraphics[width=\linewidth]{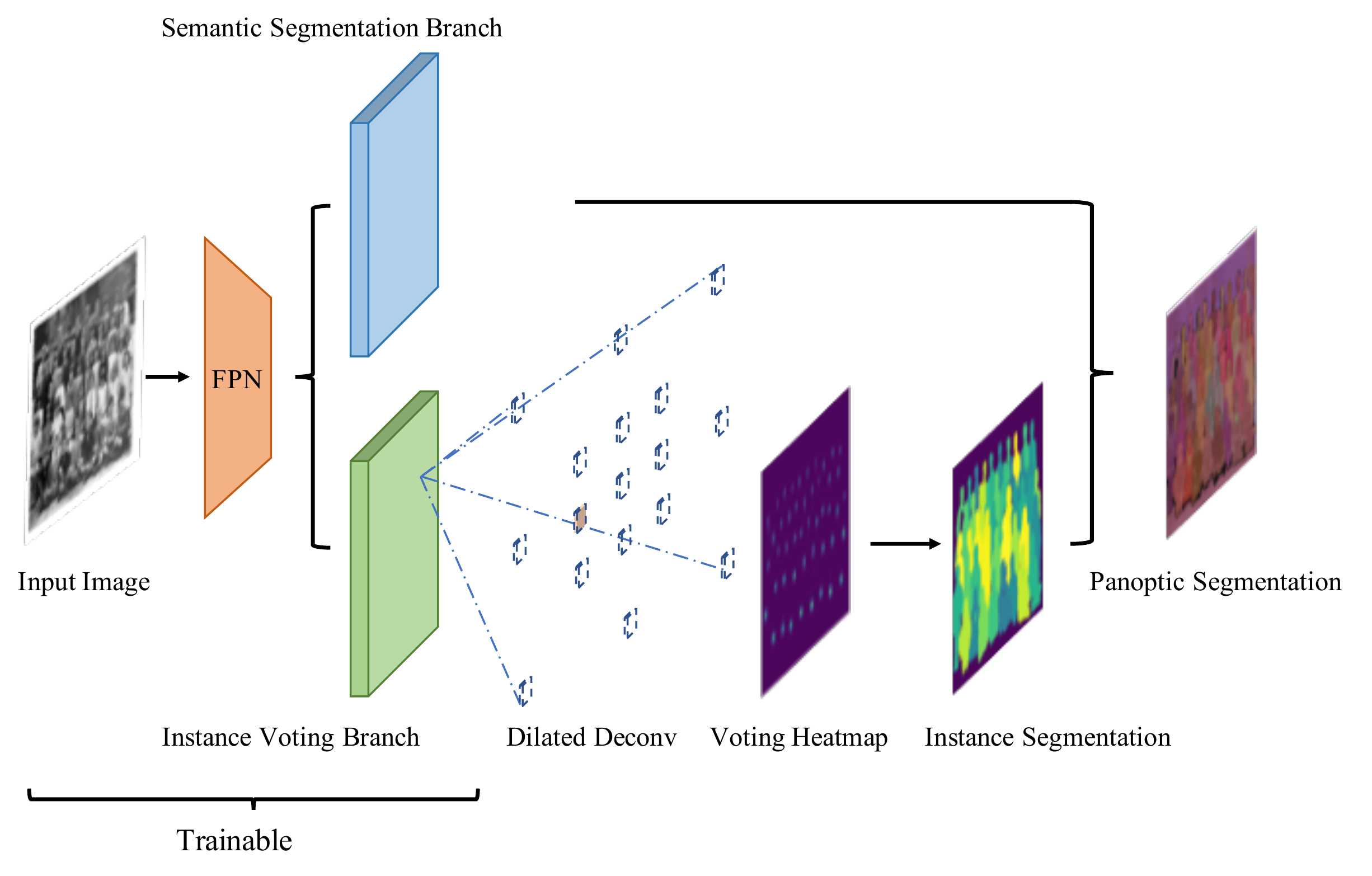}
    \vspace{-2em}
    \caption{\small
        Network architecture for PCV.
        FPN serves as a shared feature extractor for the semantic segmentation branch
        and instance voting branch. Each branch predicts output at
        every pixel, and is trained with a per-pixel cross entropy loss.
    }
    \label{fig:architecture}
\end{figure}

Given an input image, PCV starts with a convolutional neural network extracting
a shared representation (feature tensor) that is fed to two independent
sub-branches (Fig.~\ref{fig:architecture}). The semantic segmentation branch
predicts the category label for every pixel. The instance voting branch
predicts for every pixel whether the pixel is part of an instance mask, and if
so, the relative location of the instance mask centroid. This prediction is
framed as classification over a set of grid cells around a pixel according to
the \vf. Both branches are trained with standard cross-entropy loss.

The predictions from the voting branch are aggregated into a voting heatmap
(\emph{accumulator array} in the Hough transform terminology). A key technical
innovation of PCV is a dilated convolutional mechanism that implements this
efficiently. Local maxima of the heatmap are detection candidates. At each \pr,
we convolve a \qf to backproject the pixels that favor this particular peak
above all others. These pixels together form a category-agnostic instance
segmentation mask. Finally, we merge the instance and semantic segmentation
masks using a simple greedy strategy, yielding a complete panoptic
segmentation output.

\subsection{Backbone and Feature Extraction}
Our work develops a meta architecture to model and segment instances. To this
end, PCV reduces the training of instance recognition to pixel labelling,
which can be tackled by various implementations of Fully Convolutional Networks
\cite{FCN}. We follow the design of UPSNet~\cite{upsnet}, which repurposes
a Feature Pyramid Network (FPN)~\cite{FPN} with a ResNet
backbone~\cite{he2016deep} for semantic segmentation. Features from each stage
of FPN, respectively at $1/32, 1/16, 1/8$ and $1/4$ of input resolution,
first go through a shared deformable convolution module before being upsampled
to a uniform size of $1/4$ of the input scale. Channel dimensions of the
feature maps are reduced from $256$ to $128$ with $1\times1$ conv before
channel-wise concatenation. On top of this, we apply a $1\times1$ conv, softmax
and $4\times$ nearest neighbor upsampling to generate per-pixel labels. Note
that we apply softmax first, before upsampling, since it is faster to produce
instance masks at a lower resolution. The semantic segmentation branch predicts
the labels for all categories, and is different from PFPN \cite{PFPN}, which
lumps all ``thing'' classes into a single category.

\subsection{Region Discretization}
\label{sec:32}
Consider an instance mask consisting of a set of pixels
$\{ p_i \,|\, p_i \in \mathbb{R}^2 \}_{i=1}^N$,
and the instance center of mass $c = \frac{1}{N} \sum p_i$.
Predicting the relative offset $\delta_i = c \,-\, p_i$ from a pixel is
typically treated as offset regression \cite{liang2017proposal,
  neven_clustering, deeperlab, semiconv}.
But a direct regression limits the ability of the system to represent
uncertainty, and suffers from the typical problem of ``regressing to the
mean''. A pixel unsure about its instance centroid location might hedge by
pointing between multiple candidates, creating spurious peaks and false
positives. Moreover, it is impossible to attribute such pixels to object
hypotheses during backprojection. We instead frame voting as classification
among possible spatial cells where the centroid may reside. The probability
histogram produces an explicit distribution for downstream reasoning.

\begin{figure}[tb!]
\centering
\includegraphics[width=\linewidth]{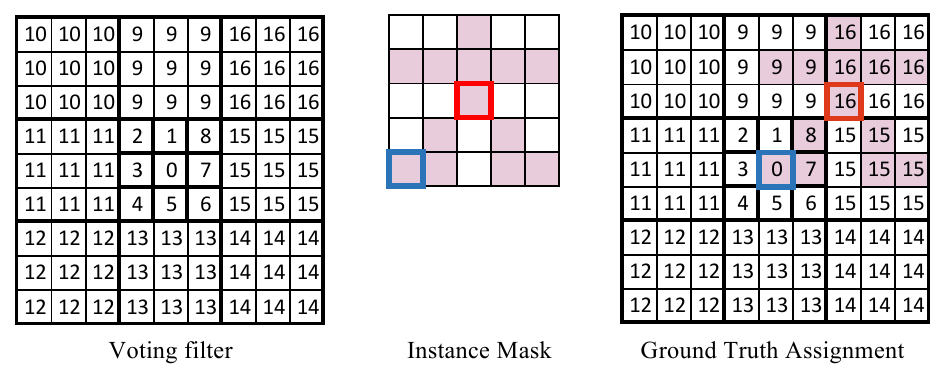}
\vspace{-1.5em}
\captionof{figure}{\small
    \vf and ground truth assignment.
    \textbf{Left}: a toy \vf mapping a $M \times M,\, M = 9$ region around a pixel to
    $K = 17$ indices. The discretization is coarser on the periphery with
    $3 \times 3$ cells.
    \textbf{Middle}: an instance mask where the red pixel is the mask centroid. We
    need to discretize the offset from the blue pixel to the centroid.
    \textbf{Right}: overlaying the \vf on top of the blue pixel, one sees that
    the ground truth voting index for the blue pixel is $16$.
}
\label{fig:voting_filter}
\end{figure}

\paragraph{Voting Filter}
Unlike YOLO~\cite{yolo}, which divides up the entire image into a grid of regularly 
tiled patches, we consider a discretization of the region centered around each pixel.
See Fig.~\ref{fig:teaser} (left) for a visual illustration. For a
particular pixel $p_i$, we map each of the $M \times M$ pixels centered around
$p_i$ to $K$ discrete indices. This mapping can be naturally recorded with a
translation invariant lookup table of size $M \times M$. By overlaying the
lookup table on top of $p_i$, the ground truth index for classification can be
directly read off from the spatial cell into which the instance centroid falls.
We refer to this lookup table as the \vf. Fig.~\ref{fig:voting_filter} shows a
toy example. For stuff pixels that do not belong to any instances, we create an
``abstention" label as an extra class, and hence there are in total $K + 1$
classes for the instance voting branch. If the instance centroid falls outside
the extent of the \vf, \ie the pixel is too far away from the centroid, we
ignore it during training.

\begin{figure}[tb!]
\centering
\begin{minipage}[c]{.5\linewidth}
\includegraphics[width=.9\linewidth]{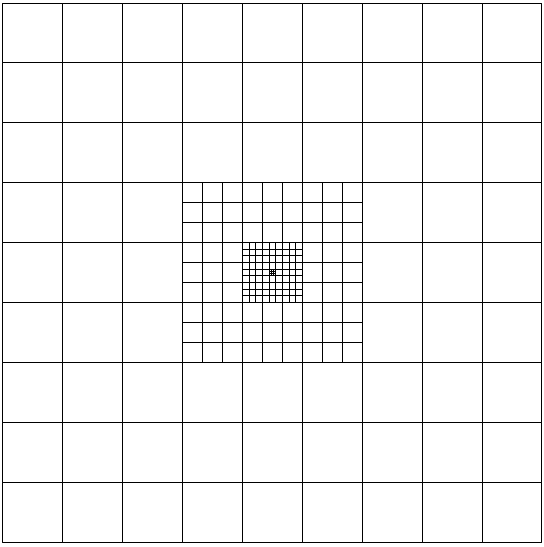}    
\end{minipage}%
\begin{minipage}[c]{.5\linewidth}
\captionof{figure}{
    The grid structure of our voting and query
    filters. It covers an area of $ 243 \times 243 $ and 
    consists of 233 bins ranging in size of $1, 3, 9, 27$ from
    center to the periphery.
}
\label{fig:grid}
\end{minipage}
\end{figure}

\begin{figure}[tb!]
  \centering
  \includegraphics[width=\linewidth]{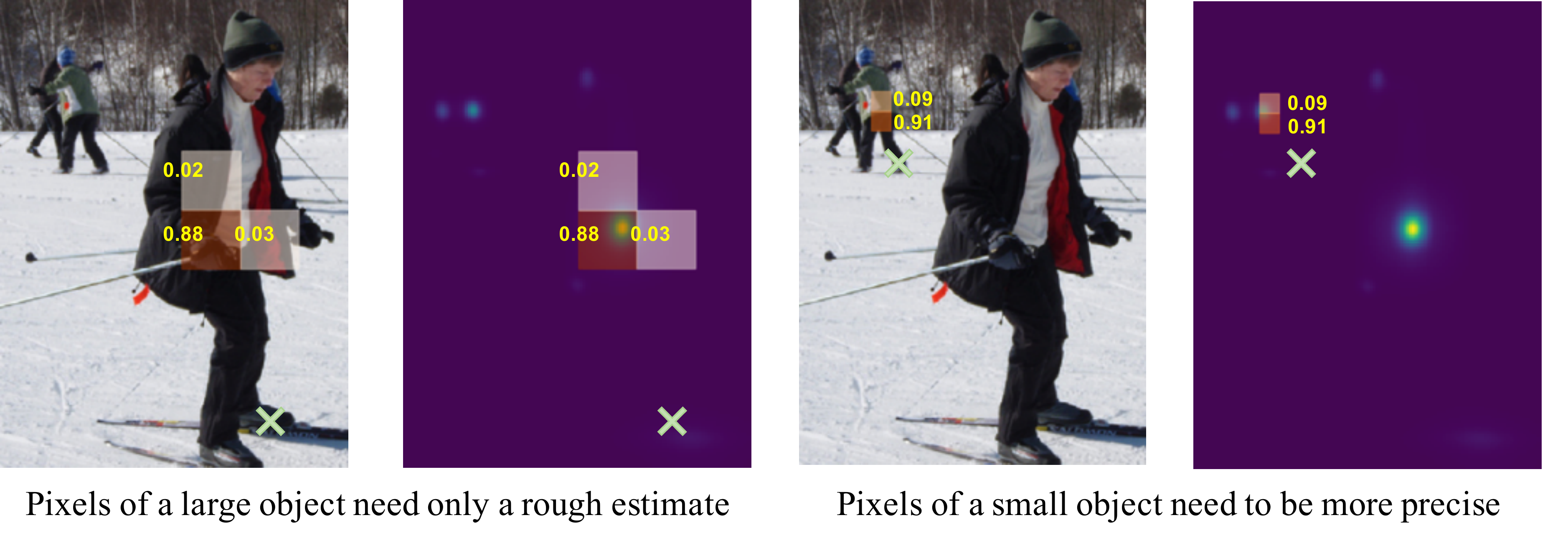}
\vspace{-1.5em}
\captionof{figure}{\small
    Illustration of voting behavior. The green cross indicates a pixel whose
    votes we inspect. We only display cells which receive a vote stronger than $ 0.01 $
    \textbf{Left}: pixels distant from the object center (like the front skier's
    foot) can afford more slack/spatial uncertainty, voting for larger
    cells near grid perimeter.
    \textbf{Right}: Pixels of small objects (the back skier's
    foot) need more spatial accuracy,
    and vote for small cells near the grid center.
  }
  \label{fig:scale_vs_acc}
\end{figure}

\paragraph{Scale \vs Accuracy}
Discretization implies a loss of spatial accuracy, but we argue that knowing
the exact location of the centroid is not necessary for accurate instance
segmentations. What matters is the consensus among pixels that enables
backprojection. Large instances can naturally tolerate coarser predictions
than small objects, as seen in Fig.~\ref{fig:scale_vs_acc}. We construct the \vf so
that the farther the distance from the instance centroid, the larger the
spatial cell. A naive evenly spaced grid would have to either be too fine,
introducing too many classes for the network to learn and converge, or too
coarse to allow accurate predictions for smaller objects. Based on these
considerations, we propose a grid of square cells whose size expands radially
outward, as shown in Fig.~\ref{fig:grid}. It involves $K = 233$ cells over
a region of $M = 243$ pixels applied to images at $1/4$ input resolution, thus
covering up to $972 \times 972$ at full resolution.

\subsection{Voting as Transposed Convolution}
The instance voting branch yields a tensor of size $[H, W, K+1]$, where $K+1$
is the number of distinct voting possibilities, including abstention by stuff
pixels. We use dilated deconvolution and average pooling to aggregate the
probabilistic votes to their intended spatial locations.

Recall our toy example in Fig.~\ref{fig:voting_filter}. Say the blue pixel
predicts a probability of $0.9$ for its instance centroid to fall into cell
$16$, which consists of $9$ pixels. Voting involves two steps: 1) transfer the
probability $0.9$ to cell $16$, and 2) share the vote evenly among the $9$
constituent pixels, with each pixel receiving $0.1$. We implement step 1 with
dilated deconvolution (deconv), and step 2 with average pooling.

Transposed convolution, or deconvolution by convention, spreads a point signal over
multiple spatial locations, whereas a conv kernel aggregates spatial information
to a single point. It is most often used during the backpropagation of a convnet, as well
as feature upsampling. The parameters of a deconv kernel can be learned. 
For the purpose of vote aggregation, however, 
we fix the deconv kernel parameters to 1-hot across each channel that
marks the target location. Dilation in this case enables a pixel to cast its
vote to faraway points. Note that for every deconv kernel, 
there exists an equivalent conv kernel, and vice versa. 
The distinction is superfluous, but thinking of and implementing voting as deconv is more natural.

Our toy \vf in Fig.~\ref{fig:voting_filter} discretizes the $9\times9$
region into inner $3 \times 3$ cells of side length $1$, encircled by outer
$3 \times 3$ cells of side length $3$, hence $K = 9 + 8 = 17$ voting classes.
At step 1, after discarding abstention votes, we split the $[H, W, 17]$ tensor
along channels into two components of size $[H, W, 9]$ and $[H, W, 8]$,
and apply two deconv kernels of size
$[C_{\text{in}} \!\! = \!\! 9, C_{\text{out}} \!\! = \!\! 1, H \!\! = \!\! 3, W \!\! = \!\! 3]$
with dilation $1$ and
$[C_{\text{in}} \!\! = \!\! 8, C_{\text{out}} \!\! = \!\! 1, H \!\! = \!\! 3, W \!\! = \!\! 3]$
with dilation $3$ to produce two heatmaps
$H_{dilate1}, H_{dilate3}$, both of size $[H, W, 1]$.

After step 1, all the votes have been sent to the center of each spatial cell.
At step 2, we smooth out the votes evenly within each cell. Smoothing in this
particular case is exactly equivalent to average pooling. We apply $3\times3$
average pooling on $H_{dilate3}$, and $1 \times 1$ average pooling on
$H_{dilate1}$ (an identity operation). The two heatmaps are then summed
together to complete the final voting heatmap. The voting process for other
instantiations of \vf can be done analogously. Voting with our default grid design 
takes on average 1.3 ms over COCO images, show in Table \ref{tab:timing}.

\begin{figure*}[ht]
\begin{minipage}[c]{.65\linewidth}
\includegraphics[width=.32\textwidth]{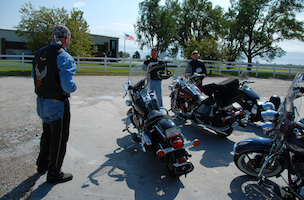}
\includegraphics[width=.32\textwidth]{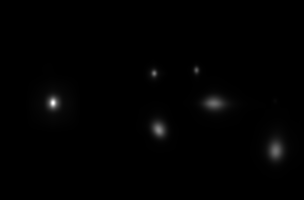}
\includegraphics[width=.32\textwidth]{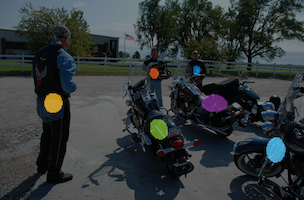}
\end{minipage}%
\begin{minipage}[c]{.35\linewidth}
\includegraphics[width=.32\textwidth]{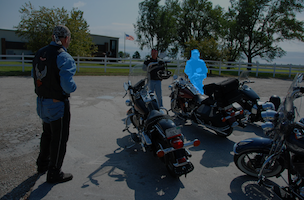}
\includegraphics[width=.32\textwidth]{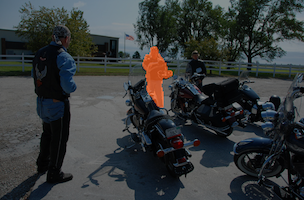}
\includegraphics[width=.32\textwidth]{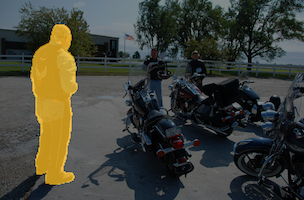}  \\
\includegraphics[width=.32\textwidth]{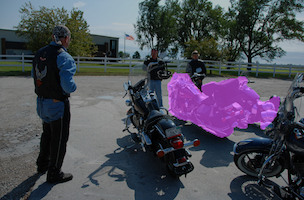}
\includegraphics[width=.32\textwidth]{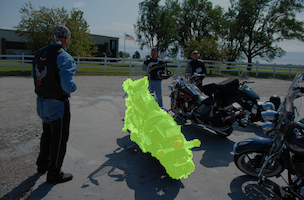}
\includegraphics[width=.32\textwidth]{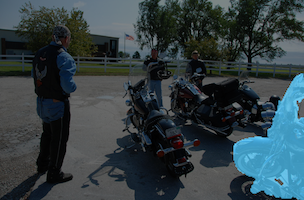}
\end{minipage}
\vspace{-1em}
\captionof{figure}{
    \small Illustration of instance mask inference in PCV. 
    Left to right: input image, voting heatmap, detected peak regions (random
    colors assigned to each peak); six masks resulting from
    backprojection from color-matching regions.
}
\label{fig:insfig}
\end{figure*}

Peaks in the voting heatmap correspond to consensus detections, and we use a
simple strategy of thresholding followed by connected components to locate the
peaks. We define a \pr, which identifies a hypothesized instance, as a
connected component of pixels that survive after thresholding the voting
heatmap. We set the threshold value to $4.0$ for both COCO and
Cityscapes. See Fig.~\ref{fig:insfig}.

\begin{figure}[!bt]
\centering
\includegraphics[width=1\linewidth]{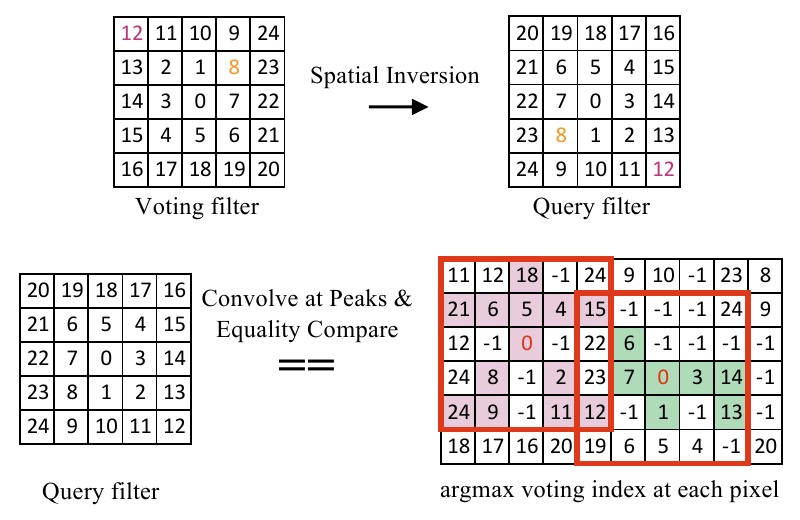}
\vspace{-2em}
\captionof{figure}{\small 
    \qf and backprojection. 
    \textbf{Top}: The \qf is a spatial inversion of the \vf, and the indices of the two filters
    are symmetric about the center (highlighted here for a few
    corresponding pairs of cells in the two filters).
    The \vf captures the spatial relationship between
    a pixel and the surrounding centroid, whereas the \qf represents
    the dual relationship between a centroid and surrounding pixels; 
    \textbf{Bottom}: The \qf is convolved within each \pr to produce instance masks. For 
    simplicity a \pr here is a single red point, but is in general a connected
    component of pixels, and hence the need for convolving the \qf. $ -1 $ denotes
    pixels whose $ \argmax $ voting decision is abstention \ie
    ``stuff" pixels. Where the \qf disagrees with the $\argmax$ voting indices,
    the pixels are not included in the instance masks.
}
\label{fig:backproj}
\end{figure}

\subsection{Backprojection as Filtering}
Backprojection aims to determine for every \pr the pixels that favor this
particular maximum above all others. To do this we make use of the \qf.
Recall that the \vf records the class label a pixel at the center of
the filter should predict given possible centroid locations around. 
The \qf is the spatial inversion of the \vf. It records the
class labels the \emph{surrounding} pixels should predict for the instance
centroid at the center of the filter. This dual relationship is
shown in the top row of Fig.~\ref{fig:backproj}.

During backprojection, we first obtain the $\argmax$ voting index at each
pixel. This is a tensor of size $[H, W, 1]$. Then, within a \pr, we convolve
the \qf and perform equality comparison against the argmax voting indices to
pick up all the pixels whose strongest vote falls within this \pr. See
Fig.~\ref{fig:backproj}, bottom row. This operation is parallelizable and
can be implemented to run on a GPU. In practice, we extend the equality
comparison to top 3 votes rather than just the $\argmax$ vote, so that a pixel
whose $\argmax$ decision is wrong is not completely abandoned. If a single
pixel is contested by multiple peaks, the pixel is assigned to the \pr whose
total vote count is the highest. In the edge case where multiple peaks are
within the same spatial cell with respect to the pixel, the pixel goes to the
spatially nearest peak (this distance is measured from the center of the
enclosing bounding box of the \pr to the pixel).

\subsection{Segment Loss Normalization}
Training a network to solve pixel labeling problems such as ours usually
involves averaging the per-pixel cross entropy loss over an image~\cite{FCN}.
Each pixel contributes equally to the training and the notion of instance is
absent. This is often the case for semantic segmentation, since the annotations
specify only the categories. For Panoptic Segmentation, however, each instance
segment is given equal weighting during evaluation and training with the default
pixel-averaged loss would put emphasis primarily on large instances, neglecting
small objects that are numerous and critical.
Therefore, we need to design an objective function that balances the loss
across instances. Let $a_i$ denotes the area of the mask segment to which a
pixel $p_i$ belongs. The training losses for both semantic and voting branches
are normalized to be
\begin{align}
  L = \frac{1}{\sum_i{w_i}}\sum_i w_i \log p(y_i|p_i)
\end{align}
where $y_i$ is the ground truth semantic/voting label, and
$w_i = \frac{1}{a_i^\lambda}$.
$\lambda$ controls the strength of normalization. When $\lambda = 0$, $w_i = 1$,
we get back the default pixel-averaged loss. When $\lambda = 1$, we divide the
summed loss from each segment by the segment area, so that all segments would
contribute equally to the training. $\lambda = 0.5$ could be interpreted as a
length based normalization that strikes a middle ground between pixel-averaged
loss and full segment normalized loss. Note that stuff and thing segments are
treated identically. The final loss is the sum of semantic segmentation loss
and voting loss
$L_{\text{total}} = L_{\text{sem}} + L_{\text{vote}}$.
In Sec.~\ref{sec:exps}, we demonstrate through ablation experiments that
segment loss normalization significantly improves performance on both COCO and
Cityscapes.

\subsection{Determining Object Categories}
Once an instance mask is obtained from backprojection, we predict its category
by taking the majority decision made by the semantic segmentation branch in the
masked region. This strategy is similar to the one used by \cite{deeperlab}.

%% file: exps.tex
\begin{table}[b!]
\vspace{-3mm}
\tablestyle{6pt}{1.2}
\begin{tabular}{x{30} c |x{8} x{8} x{8} x{8} x{8} x{8} x{8} x{8} x{8}}
  &  & PQ  & SQ  & RQ  & PQ$^{th}$ & SQ$^{th}$ & RQ$^{th}$ & PQ$^{st}$\\
\shline
\multirow{4}[2]{*}{COCO} & 1/4 gt
& \textbf{92.5} & 93.2 & \textbf{99.2} & \textbf{90.6} & 91.6 & \textbf{98.8} & 95.3 \\
& default grid & 90.1 & 93.0 & 96.8 & 86.6 & 91.3 & 94.8 & 95.3 \\
& simple grid  & 79.2 & 92.6 & 85.1 & 68.6 & 90.7 & 75.4 & 95.3 \\
& uniform grid & 67.1 & \textbf{95.8} & 70.1 & 49.4 & \textbf{96.0} & 51.5 & 93.8 \\
\shline
\multirow{4}[2]{*}{Cityscapes} & 1/4 gt
& \textbf{89.4} & 89.8 & \textbf{99.6} & \textbf{87.1} & 87.7 & \textbf{99.4} & 91.0  \\
& default grid & 88.6 & 89.7 & 98.8 & 85.4 & 87.4 & 97.6 & 91.0 \\
& simple grid  & 83.0 & 89.3 & 92.8 & 72.1 & 86.6 & 83.4 & 91.0 \\
& uniform grid & 66.1 & \textbf{92.6} & 71.8 & 31.8 & \textbf{94.3} & 33.4 & 91.0 \\
\end{tabular}
\vspace{-1mm}
\caption{
   Oracle inference on COCO and Cityscapes \texttt{val} using ground
   truth voting and semantic classification label. `1/4 gt' is the performance
   upper bound when the output is at 1/4 of input resolution, and the default
   discretization is behind by a small gap.
}\label{tab:oracle}
\vspace{-3mm}
\end{table}

\begin{table*}[t]
\centering

\subfloat[
    \small\textbf{Segment Loss Normalization:} results on COCO \texttt{val}
    with various $\lambda$ that controls the normalization strength. $\lambda = 0.5$
    improves PQ$^{th}$ by 7 points over the commonly used pixel-averaged loss.
    \label{tab:loss_norm_abla}
]
{\makebox[0.475\linewidth][c]{
\tablestyle{3pt}{1.1}
\begin{tabular}{l|x{20}x{20}x{20}|x{20}x{20}x{20}|x{20}}
 & PQ & SQ & RQ & PQ$^{th}$ & SQ$^{th}$ & RQ$^{th}$ & PQ$^{st}$ \\
\shline
$\lambda$=0 & 32.9 & 77.2 & 40.5 & 33.0 & 78.0 & 40.3 & 32.6 \\
$\lambda$=0.5 & \textbf{37.5} & \textbf{77.7} & \textbf{47.2} & \textbf{40.0} & \textbf{78.4} & \textbf{50.0} & \textbf{33.7} \\
$\lambda$=1.0 & 31.8 & 74.7 & 41.3 & 33.9 & 75.1 & 44.0 & 28.6 \\
\end{tabular}
}}
\hspace{0.02\linewidth}
\subfloat[
    \textbf{Impact of Discretization:} consistent with oracle results
    in Table~\ref{tab:oracle}, the simple grid is too coarse for accurate
    localizations and the default grid leads on PQ$^{th}$ by 7.17 points on
    COCO \texttt{val}.
    \label{tab:grid_abla}%
]
{\makebox[0.475\linewidth][c]{
\tablestyle{3pt}{1.1}
\begin{tabular}{l x{20}x{20}x{20}|x{20}x{20}x{20}|x{20}}
& PQ & SQ & RQ & PQ$^{th}$ & SQ$^{th}$ & RQ$^{th}$ & PQ$^{st}$ \\
\shline
default grid & 37.5 & 77.7 & 47.2 & 40.0 & 78.4 & 50.0 & 33.7  \\
simple grid & 33.3 & 77.2 & 41.8 & 32.8 & 77.4 & 40.9 & 34.1  \\
\end{tabular}
}}
\vspace{-2mm}
\caption{
    Ablations on segment loss normalization and impact of discretization.
    \label{tab:ablation}
}
\vspace{-2mm}
\end{table*}

\begin{table*}[t!]
\tablestyle{5pt}{1.1}
\begin{tabular}{clcc| x{20}x{20}x{20} | x{20}x{20}x{20} | x{20}x{20}x{20} }
& Methods & Backbone & Split & PQ & SQ & RQ & PQ$^{th}$ & SQ$^{th}$ & RQ$^{th}$ & PQ$^{st}$ & SQ$^{st}$ & RQ$^{st}$ \\
\shline
\multirow{3}[2]{*}{Mask R-CNN} & PFPN~\cite{PFPN} (1x) & ResNet 50 & val & 39.4 & 77.8 & 48.3 & 45.9 & 80.9 & 55.4 & 29.6 & 73.3 & 37.7 \\
& PFPN~\cite{PFPN} (3x) & ResNet 50 & val & 41.5 & 79.1 & 50.5 & 48.3 & 82.2 & 57.9 & 31.2 & 74.4 & 39.4 \\
& UPSNet~\cite{upsnet} (1x) & ResNet 50 & val & 42.5 & 78.0 & 52.5 & 48.6 & 79.4 & 59.6 & 33.4 & 75.9 & 41.7 \\
\shline
\multirow{2}[2]{*}{Single Stage Detection} & SSPS~\cite{SSPS} & ResNet 50 & val & 32.4 & - & - & 34.8 & - & - & 28.6 & - & - \\
& SSPS~\cite{SSPS} & ResNet 50 & test-dev & 32.6 & 74.3 & 42.0 & 35.0 & 74.8 & 44.8 & 29.0 & 73.6 & 37.7 \\
\shline
\multirow{7}[1]{*}{Proposal-Free} & AdaptIS~\cite{adapis} & ResNet 50 & val & 35.9 & - & - & 40.3 & - & - & 29.3 & - & - \\
& DeeperLab~\cite{deeperlab} & Xception 71 & val & 33.8 & - & - & - & - & - & - & - & - \\
& DeeperLab~\cite{deeperlab} & Xception 71 & test-dev & 34.3 & 77.1 & 43.1 & 37.5 & 77.5 & 46.8 & 29.6 & 76.4 & 37.4 \\
& SSAP~\cite{ssap} & ResNet 101 & val & 36.5 & - & - & - & - & - & - & - & - \\
& SSAP~\cite{ssap} & ResNet 101 & test-dev & 36.9 & 80.7 & 44.8 & 40.1 & 81.6 & 48.5 & 32.0 & 79.4 & 39.3 \\
\cmidrule{2-13}
& \textbf{Ours} (1x) & ResNet 50 & val & \textbf{37.5} & 77.7 & \textbf{47.2} & 40.0 & 78.4 & \textbf{50.0} & \textbf{33.7} & 76.5 & \textbf{42.9} \\
& \textbf{Ours} (1x) & ResNet 50 & test-dev & \textbf{37.7} & 77.8 & \textbf{47.3} & \textbf{40.7} & 78.7 & \textbf{50.7} & \textbf{33.1} & 76.3 & \textbf{42.0} \\
\end{tabular}
\vspace{-1em}
\caption{
    \small Comparisons on COCO. PCV outperforms proposal-free
    and single-state detection methods.
}
\label{tab:coco}
\end{table*}

\begin{table}
\tablestyle{5pt}{1.1}
\begin{tabular}{l x{25}x{25}x{25}x{25} }
 & PQ & PQ$^{th}$ & PQ$^{st}$ & mIoU \\
\shline
DIN~\cite{DIN}  & 53.8  & 42.5  & 62.1  & 80.1 \\
UPSNet~\cite{upsnet} & 59.3  & 54.6  & 62.7  & 75.2 \\
PFPN~\cite{PFPN}  & 58.1  & 52.0  & 62.5  & 75.7 \\
AdaptIS~\cite{adapis} & 59.0  & 55.8  & 61.3  & 75.3 \\
SSAP~\cite{ssap}  & 58.4  & 50.6  & -     & - \\
\textbf{Ours} & 54.2 & 47.8 & 58.9 & 74.1\\
\end{tabular}
\caption{
    Results on Cityscapes val using ResNet 50
}
\label{tab:cityscapes}
\end{table}

\begin{table}
\tablestyle{5pt}{1.1}
\begin{tabular}{x{28} x{42} x{25} x{25} x{25} x{25}}
& Input Size & Backbone & Voting & Backproj. & Total \\
\shline
COCO       &  $~~ 800 \times 1333$  & 93.4  & 1.3   & 81.8  & 176.5 \\
Cityscapes & $1024 \times 2048$ & 115.6 & 2.8   & 64.4  & 182.8 \\
\end{tabular}
\caption{
    runtime benchmark using GTX 1080 Ti (unit: ms)
}
\label{tab:timing}
\end{table}

\section{Experiments}
\label{sec:exps}
We report results on COCO \cite{mscoco} and Cityscapes \cite{cityscapes}
Panoptic Segmentation. Since PCV formulates centroid prediction as
region classification rather than offset regression, it trades off the upper
bound on prediction accuracy for a richer representation. We first conduct
oracle experiments to understand the potential of our system. Then we compare our
model performance on COCO and Cityscapes validation sets against prior and concurrent
works. Ablations focus on the use of different discretization schemes and segment loss
normalizations.

\subsection{Setup}
COCO Panoptic Segmentation consists of 80 thing and 53 stuff categories.
We use the 2017 split with 118k training images and report results on \texttt{val}
and \texttt{test-dev} splits. Cityscapes includes images of urban street scenes.
We use standard \texttt{train}/\texttt{val} split,
including 2975 and 500 images respectively. There are 19 categories, 11 stuff and 8 things.
We measure performance by Panoptic Quality (PQ)~\cite{PStask}.
PQ can be interpreted as a generalized F1-score that reflects both
recognition quality RQ and segmentation quality SQ.
In addition to the overall PQ, we include PQ$^{th}$
and PQ$^{st}$ and focus in particular on thing category performance.

\subsection{Oracles}
There are no learnable parameters in the vote aggregation and backprojection steps of PCV,
and so once the pixel-wise classification decisions are made by the backbone network,
the subsequent inference is deterministic. Therefore we perform oracle experiments
by feeding into the inference pipeline ground truth
classification labels for both the voting and semantic branches. As seen in Table
\ref{tab:oracle}, given our default discretization scheme, PCV oracle achieves performance close to the
upper bound on both COCO and Cityscapes validation sets. The remaining gaps in PQ are mostly
due to small instances of extremely high occlusion and instances with colliding centroids.
We also show 2 more oracle results: a simple radially expanding grid with 41 voting
classes performs worse than the default grid with 233 voting classes. A uniform grid
with evenly-spaced bins of size 15 and total \vf side length 225 does the worst.
Even though it has roughly the same number of voting classes as our default grid,
the even spacing severely degrades performance on small instances.

\subsection{Main Results and Ablations}
For Cityscapes training we use batch size 16 over 8 GPUs and crop input images to
a uniform size of $1536 \times 1024$. We apply random horizontal flipping and randomly
scale the size of the crops from 0.5 to 2. The model is trained for 65 epochs~(12k iterations)
with learning rate set initially at 0.01, dropped by 10x at 9000 iterations.
We use SGD with momentum 0.9 and weight decay at 1e-4.

For COCO, we use standard Mask R-CNN $ 1\times $ training schedule and hyperparameters.
Input images are resized to have length 800 on the shorter side and length
not exceeding 1333 on the longer.
Resizing is consistent for both training and testing. Left-right flipping is
the only data augmentation used. We use SGD with momentum 0.9 and set the initial learning
rate at $ 0.0025 $, weight decay at $ 0.0001 $. The model is trained on 8 GPUs with
batch size of 16 for a total of around 13 epochs (90k iterations). Learning rate
decays by a factor of 10 at 60k and 80k iterations. BatchNorm~\cite{ioffe2015batch} layers in ResNet
are frozen in our current setup.

Following \cite{PFPN, upsnet}, for stuff predictions we filter out small predicted segments to reduce
false positives. The thresholds are set at areas of 4096 pixels for COCO and 2048 pixels for Cityscapes.

\begin{figure*}[!t]
\includegraphics[width=.245\textwidth]{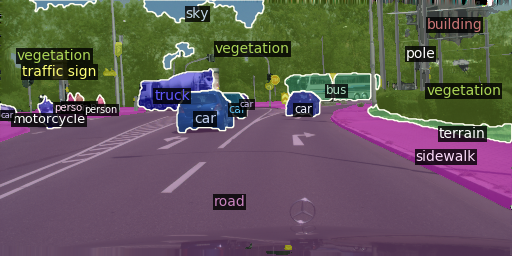}
\includegraphics[width=.245\textwidth]{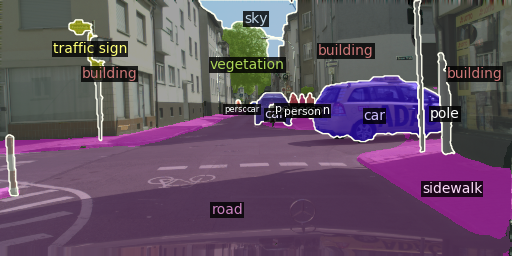}
\includegraphics[width=.245\textwidth]{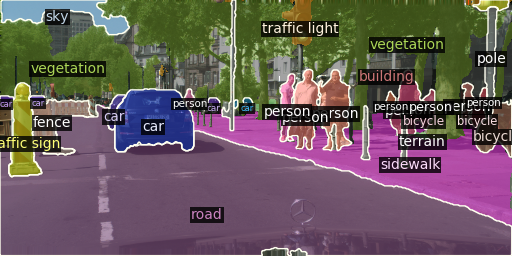}
\includegraphics[width=.245\textwidth]{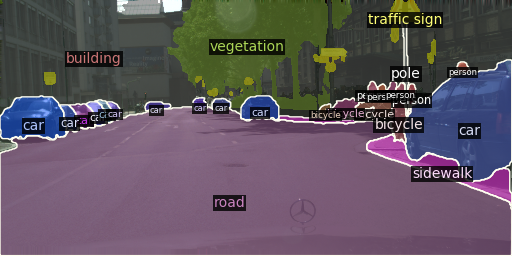}\\
\includegraphics[width=0.995\textwidth]{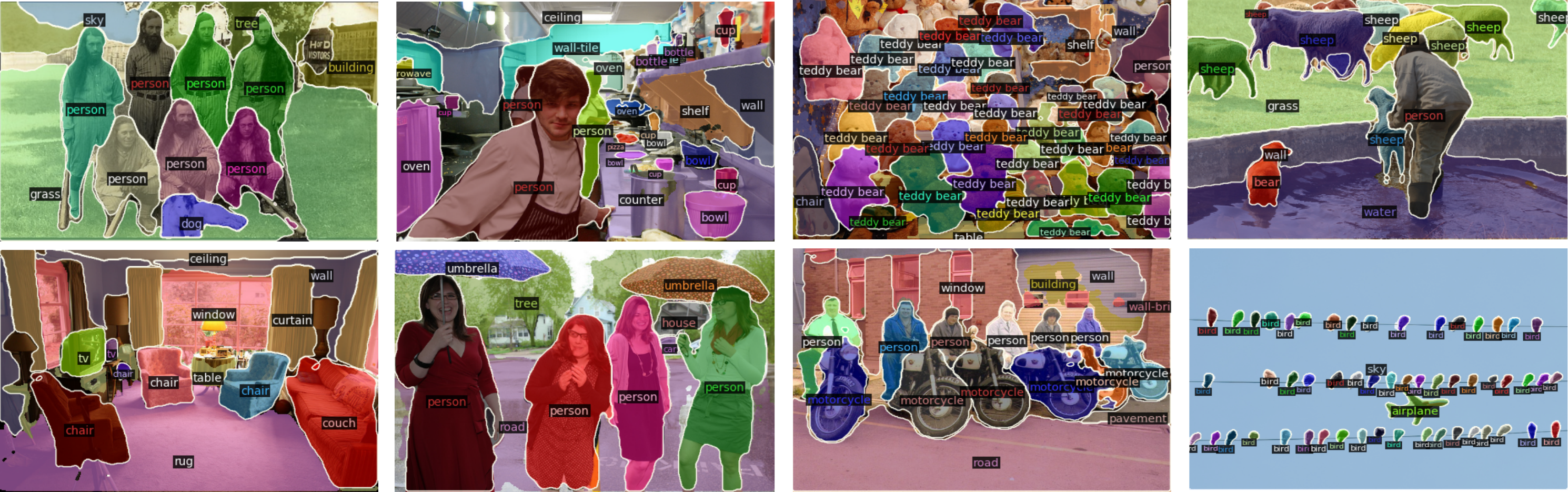}
\vspace{-.8em}
  \captionof{figure}{Results of PCV on Cityscapes \texttt{val} and COCO \texttt{test-dev}.}
  \label{fig:examples}
\end{figure*}

\begin{figure*}[!t]
\vspace{-.5em}
\includegraphics[height=.97in]{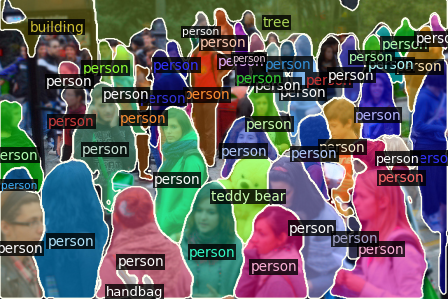}
\includegraphics[height=.97in]{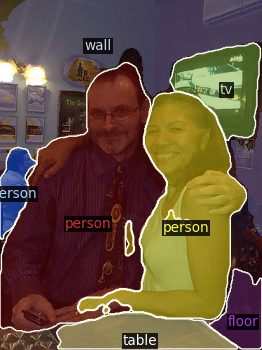}
\includegraphics[height=.97in]{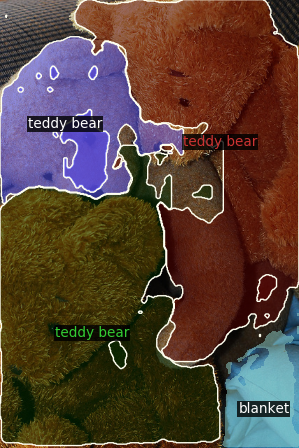}
\includegraphics[height=.97in]{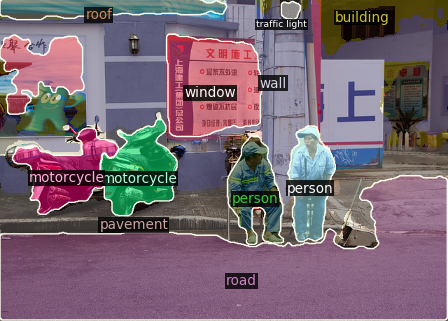}
\includegraphics[height=.97in]{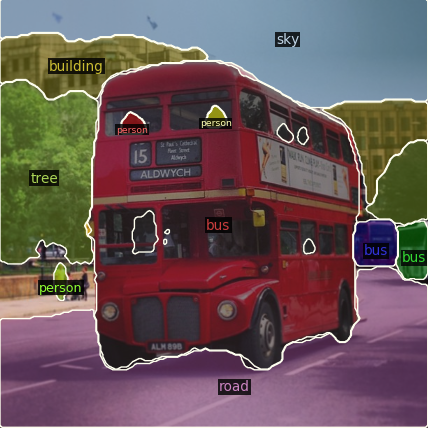}
\includegraphics[height=.97in]{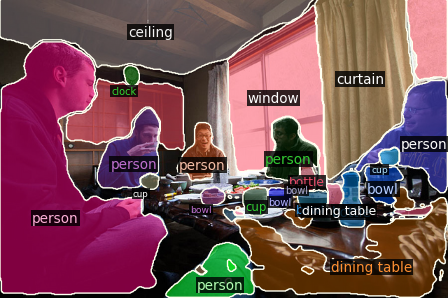}
\vspace{-.8em}
\captionof{figure}{
    Results of PCV on images from COCO ~\texttt{val2017}.
}
\label{fig:coco_val_qualititative}
\end{figure*}

\paragraph{Main Results} We compare the performance of PCV against representative
methods using different approaches. On both COCO and Cityscapes, PCV still lags
behind leading methods that leverage Mask RCNN for instance segmentation. On the challenging
COCO benchmark, PCV outperforms all other proposal free methods as well as \cite{SSPS} which uses
RetinaNet for object detection. Results on Cityscapes are shown in Table~\ref{tab:cityscapes}.
Qualitative results are displayed in Fig \ref{fig:examples} and \ref{fig:coco_val_qualititative}.

\paragraph{Ablation: discretization} We explore the influence of discretization by comparing
a model using a simple 41-cell grid against the default model using a 233-cell grid.
The results on COCO \texttt{val2017} are presented in Table~\ref{tab:grid_abla}.
The full grid outperforms the simplistic grid and this agrees with our observation
made earlier for the oracle experiments. The simple grid might make the learning easier
but sacrifices the prediction accuracy due to coarse discretization.

\paragraph{Ablation: segment loss normalization} We hypothesize that the contribution from each pixel to the
final training loss should be normalized by a function of the segment area so that large instances would not eclipse
the attention paid to small objects. We train PCV on COCO with $\lambda$ set at $0, 0.5, 1$. As expected, pixel-averaged
loss with $\lambda = 0$ dilutes the focus on small objects and drags down PQ things, while a
full area based segment normalization with $\lambda = 1 $ causes severe degradation on stuff PQ.
Length-based normalization with $\lambda $ set at 0.5 achieves the best performance on both things and stuff.

\vspace{-.5em}

\paragraph{Timing} Table \ref{tab:timing} examines PCV runtime breakdown, benchmarked
on a GTX 1080 Ti and averaged over Cityscapes and COCO \texttt{val}. Backprojection
relies on an unoptimized indexing implementation. PCV runs at 5fps.

%% file: conclusion.tex
\section{Conclusion}\label{sec:conclusions}

We propose a novel approach to panoptic segmentation that is entirely
pixel-driven. Different from proposal-based object detection approaches, 
Pixel Consensus Voting elevates pixels to a first-class role; 
each pixel provides evidence for presence and
location of object(s) it may belong to. It affords efficient inference, thanks
to our convolutional mechanism for voting and backprojection. It is significantly simpler than
current, highly engineered state-of-the-art panoptic segmentation models.

Our results demonstrate that the Generalized Hough transform, a historical
competitor to the sliding window detection paradigm, is again viable once
combined with deep neural networks. This should be a call for future research
exploring new ways of interconnecting traditional computer vision techniques
with deep learning. For PCV specifically, there is clear potential to explore
improved voting and inference protocols. This includes voting in higher
dimensions (\emph{e.g.}, scale-space) and alternative models of interaction
between instance detection and category assignments.